\documentclass[letterpaper]{article} 
\usepackage{aaai25}  
\usepackage{times}  
\usepackage{helvet}  
\usepackage{courier}  
\usepackage[hyphens]{url}  
\usepackage{graphicx} 
\usepackage{natbib}  
\usepackage{caption} 
\usepackage{algorithm}
\usepackage{algorithmic}

\usepackage{newfloat}
\usepackage{listings}
\usepackage{subfigure}
\usepackage{bm}
\usepackage{amsmath}
\usepackage{amssymb}
\usepackage{booktabs}
\usepackage{makecell}
\usepackage{multirow}
\usepackage{xcolor}

\newcommand{\ie}{\textit{i}.\textit{e}., }

\newcommand{\wrt}{\textit{w}.\textit{r}.\textit{t}. }

\newcommand{\tensor}[1]{\bm{\mathcal{#1}}}

\DeclareCaptionStyle{ruled}{labelfont=normalfont,labelsep=colon,strut=off} 
\floatstyle{ruled}
\newfloat{listing}{tb}{lst}{}
\floatname{listing}{Listing}
\title{TPCH: Tensor-interacted Projection and Cooperative Hashing for\\ Multi-view Clustering}
\author{
    Zhongwen Wang\textsuperscript{\rm 1},
    Xingfeng Li\textsuperscript{\rm 1}\thanks{Corresponding authors.},
    Yinghui Sun\textsuperscript{\rm 2}\thanks{Corresponding authors.},
   Quansen Sun\textsuperscript{\rm 1},\\
    Yuan Sun\textsuperscript{\rm 3},
    Han Ling\textsuperscript{\rm 1},
    Jian Dai\textsuperscript{\rm 4},
    Zhenwen Ren\textsuperscript{\rm 5}
}
\affiliations{
    \textsuperscript{\rm 1}School of Computer Science and Engineering, Nanjing University of Science and Technology\\
    \textsuperscript{\rm 2}	School of Computer Science and Engineering, Southeast University\\
    \textsuperscript{\rm 3}	College of Computer Science, Sichuan University\\
    \textsuperscript{\rm 4}	Southwest Automation Research Institute, China South Industries Group Corporation\\
    \textsuperscript{\rm 5}	School of National Defence Science and Technology, Southwest University of Science and Technology\\

    jankinwang@njust.edu.cn, lixingfeng@njust.edu.cn, sunyh@seu.edu.cn, sunquansen@njust.edu.cn, sunyuan$\_$work@163.com, 321106010190@njust.edu.cn, daijian1000@163.com, rzw@njust.edu.cn
}

\usepackage{bibentry}

\begin{document}

\maketitle

\begin{abstract}
In recent years, anchor and hash-based multi-view clustering methods have gained attention for their efficiency and simplicity in handling large-scale data. However, existing methods often overlook the interactions among multi-view data and higher-order cooperative relationships during projection, negatively impacting the quality of hash representation in low-dimensional spaces, clustering performance, and sensitivity to noise. To address this issue, we propose a novel approach named Tensor-Interacted Projection and Cooperative Hashing for Multi-View Clustering(TPCH). TPCH stacks multiple projection matrices into a tensor, taking into account the synergies and communications during the projection process. By capturing higher-order multi-view information through dual projection and Hamming space, TPCH employs an enhanced tensor nuclear norm to learn more compact and distinguishable hash representations, promoting communication within and between views. Experimental results demonstrate that this refined method significantly outperforms state-of-the-art methods in clustering on five large-scale multi-view datasets. Moreover, in terms of CPU time, TPCH achieves substantial acceleration compared to the most advanced current methods. The code is available at \textcolor{red}{\url{https://github.com/jankin-wang/TPCH}}.
\end{abstract}

\section{Introduction}
In recent years, advanced information-gathering technologies allow us to obtain multi-view data from the same object, enabling a more in-depth and comprehensive description of the object \cite{lu2023differentiable,qin2024TMM,sun2023hierarchicaltip,wu2023interpretable,chen2023dual,qin2024TIP,sun2023hierarchicaltmm,sun2024robust}. For example, in the field of action recognition, different views such as RGB, optical flow, and skeletal data can be used to describe actions \cite{hu_action_2023,wang_human_2022}. With the widespread application of multi-view data, a series of multi-view learning tasks have emerged, such as multi-view classification \cite{liu2024masked,han2022trusted}, multi-view feature aggregation \cite{hu2024cross,yang_robust_2020,hu_fedmmd_2024}, incomplete multi-view clustering \cite{wen2023deep,wen2023graph} and so on. Among them, Multi-View Clustering (MVC) is an unsupervised method in multi-view learning \cite{li2023auto,li2024Fast,li2024incomplete}. Compared to supervised methods, multi-view clustering does not require data to be labeled in advance, making it more efficient in explosive data growth. In practical applications, multi-view clustering has been widely used in computer vision, natural language processing, medical image analysis, and other fields \cite{fang_comprehensive_2023,wen_survey_2023,zhang_cmc_2021}. In the context of large-scale data, multi-view clustering faces challenges in computation and storage, known as the problem of Large-scale Multi-view Clustering (IMVC).

To solve the problem of large-scale multi-view clustering, early attempts involve reducing the number of matrix multiplications by seeking approximations of multi-view data matrices, thus reducing computational overhead \cite{wang_fast_nodate,zhang2016large}. In recent years, the method of selecting or generating a representative set of data from multi-view data, known as "anchors," has been widely adopted \cite{nie_fast_2022,liu_fast_2024,wang_highly-efficient_2022,zhang_large-scale_2023}. This approach significantly reduces computational complexity and memory requirements by selecting a small subset of anchor points to represent the entire dataset. Moreover, anchor-based methods establish bridges between different views through these anchor points, enhancing the exploration of consistent information between views. In large-scale clustering tasks, traditional multi-view learning methods that operate directly on the raw data incur high computational and storage costs. The emergence of Binary Multi-View Clustering (BMVC) \cite{zhang2018binary} methods has alleviated this issue. BMVC maps data to a low-dimensional binary Hamming space and performs clustering analysis in this lower-dimensional space, accelerating computational and storage efficiency in handling large-scale multi-view data. Furthermore, existing hash-based methods typically project multi-view data from the original space to a shared Hamming representation space to enhance the expressive power of the Hamming representation space.

However, existing BMVC methods typically utilize multiple independent projection matrices to project multi-view data into low-dimensional space and then learn hash codes for clustering partitioning in the low-dimensional space. The independence of projection matrices leads to insufficient communication between multi-view data and a lack of high-order collaborative relationships, hindering the learning of compact hash representations in low-dimensional space and reducing clustering performance.

To solve the problems above, we proposed a novel method called Tensor-Interacted Projection and Cooperative Hashing for Multi-View Clustering (TPCH). Specifically, multiple independent projection matrices project nonlinear multi-view data from high-dimensional space to multiple low-dimensional spaces, resulting in multiple distinctive hash representations. To enhance the high-order collaboration of projection matrices during the projection process, we first superimpose multiple projection matrices into a high-order tensor. Simultaneously, we utilize t-SVD decomposition-based enhanced tensor low-rank constraints to enhance the compactness of hash representations. In the Hamming space, multiple hash representations are stacked into high-order tensors and enhanced with high-order tensor low-rank constraints to strengthen the collaborative capabilities within and between views, further improving the compactness of hash representations across multiple views. Finally, the average of multiple compact hash representations is taken to obtain clustering results. Our work has the following main contributions:

\begin{itemize}
\item For the first time, the higher-order interactions between projection matrices are considered, and a tensor-interacted projection and cooperative hashing framework is constructed to capture the high-order multi-view information of dual projections and  Hamming space.
\item An enhanced tensor nuclear norm on the core tensor is employed to improve the compactness and distinguishability of hash codes, reducing the impact of redundant information and noise to enhance robustness.
\item Experimental results on five common large-scale multi-view datasets demonstrate that TPCH significantly improves clustering performance under five mainstream clustering evaluation metrics compared to current state-of-the-art methods.
\end{itemize}

\section{Related Work}
\subsection{Anchor-based Large-scale Multi-view Clustering}
In the process of multi-view clustering, each view provides a unique feature description for objects \cite{lu2023robust,lu2020sparse}. Anchor-based multi-view clustering methods aim to achieve more precise and robust clustering results by facilitating the exchange between views through anchors. Selecting or generating a small number of anchors from the entire dataset can significantly reduce computational complexity and memory usage, which is particularly effective when dealing with large-scale datasets. \cite{kang_large-scale_2020} learns a smaller bipartite graph for each view and then merges these bipartite graphs to obtain a unified representation. \cite{wang_highly-efficient_2022,li_parameter-free_2024} reduce complexity by learning consistent anchor points. However, this approach assumes that anchor points are shared among all views, emphasizing consistency between views while overlooking complementarity. To address this, \cite{liu_fast_2024} uses view-independent anchor points to represent data. This method allows different anchor points to be chosen for different views, providing high flexibility to adapt to the characteristics and requirements of different views. In recent years, efforts have been made to find a balance between consistency and complementarity among views, leading to the emergence of tensor anchor-based methods \cite{long2023multi,xia2022tensorized,chen2022low}. These methods stack each bipartite graph as a tensor and participate as a whole in subsequent clustering, achieving the effect of seeking consistency between views while retaining complementarity among views.

\subsection{Binary Multi-view Clustering}
Binary multi-view clustering methods analyze data by mapping it into a low-dimensional binary space, significantly reducing the dimensionality of the data, alleviating the curse of dimensionality, and improving the efficiency of clustering and subsequent processing. The low-dimensional binary hash codes enable faster similarity calculations between data points, especially when using Hamming distance, greatly enhancing clustering computation speed. Reduced-dimensional data occupies less storage space, making binary hash-based multi-view clustering methods more suitable for large-scale datasets. Binary-based clustering methods have been around since the early days \cite{gong2015web}. However, these methods separate binary representation and clustering into two separate steps, requiring customized representation methods for different datasets, thus limiting the algorithm's generalization ability. To address this issue, in recent years, researchers have adopted a joint learning approach to simultaneously learn multi-view binary representations and clustering structures, achieving good clustering results with low computational and storage costs \cite{shen2017compressed,zhao_multi-view_2023}. To obtain better binary representations, \cite{zhang2018binary} places images into a compact common binary space and uses a binary matrix decomposition model for clustering binary representations. \cite{zhang_learning_2024} employs novel tensor low-rank constraints to stack binary representations from various views into tensors to capture high-order interactions.

\section{Methodology}
To derive bipartite graphs from raw multi-view data, non-linear function methods effectively capture the inherent structure and semantic relationships within the data. These methods bring similar data closer together in high-dimensional space while pushing dissimilar data farther apart. Inspired by \cite{zhang_learning_2024}, we employ a non-linear Radial Basis Function (RBF) kernel mapping to construct bipartite graphs for each view. The corresponding formula is
\begin{equation}
\fontsize{7pt}{7pt}\selectfont
\begin{aligned}
  \phi\left( X_{p} \right) = \left\lbrack {\exp\left( {- \frac{\left| \left| {X_{p}^{1} - S_{p}^{1}} \right| \right|^{2}}{\delta}} \right)},\ldots,{\exp\left( {- \frac{\left| \left| {X_{p}^{n} - S_{p}^{m}} \right| \right|^{2}}{\delta}} \right)} \right\rbrack~^{T}
\end{aligned}
\label{eq.1}
\end{equation}
where {$\phi(\cdot)$} denotes a mapping used to obtain the bipartite graph for each view. \textbf{$\delta$} represents the width of the kernel. We randomly select \textbf{$m$} anchor point samples \textbf{$S_p$} from the p-th view to ensure that the original data structure remains intact after nonlinear mapping.

\subsection{Problem Formulation}
\begin{figure}[h]
  \centering
  \includegraphics[width=\linewidth]{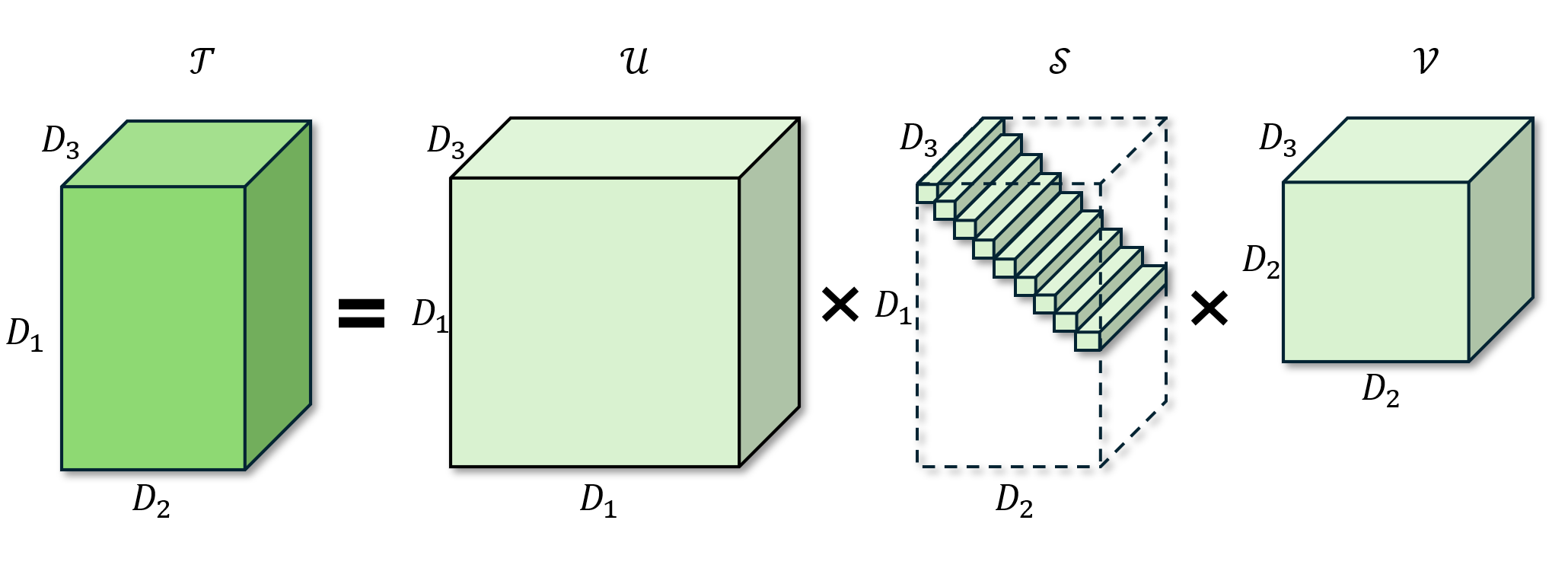}
  \caption{Basic process of t-SVD. {$ \tensor{T} \in R^{D_{1} \times D_{2} \times D_{3}}$} represents the tensor to be decomposed,  {$\tensor{S} \in R^{D_{1} \times D_{2} \times D_{3}}$} denotes the core tensor, {$\tensor{U} \in R^{D_{1} \times D_{1} \times D_{3}}$}, and {$\tensor{V} \in R^{D_{2} \times D_{2} \times D_{3}}$} represent the left and right tensors, respectively, resulting from the t-SVD decomposition of {$\tensor{T}$}. }
  \label{fig.t-svd}
\end{figure}

Based on the kernelized bipartite graphs, we propose to explore the high-order information in dual projection and Hamming space, facilitating intra-view and inter-view communication and synergy of the projection matrices and hash codes. In this way, we could learn more compact and distinguishable hash codes. Concretely, the proposed \textbf{Tensor-interacted Projection and Cooperative Hashing for Multi-view Clustering (TPCH)}  can be implemented as
\begin{equation}
\begin{aligned}
& \min_{\substack{ \mathbf{B}_p,\mathbf{Q}_p, \\ \tensor{Q},\tensor{B}}}\alpha\sum_{p=1}^{v} \|\mathbf{Q}_p^\top \phi{(\mathbf{X}_p)} - \mathbf{B}_p\|_{F}^{2}+(\|\tensor{Q}\|_{tnn}+ \|\tensor{B}\|_{tnn})  \\& \text{s.t.}\ \tensor{B} =\Phi([\mathbf{B}_1 ; \cdots ; \mathbf{B}_v]),\tensor{Q} =\Phi([\mathbf{Q}_1 ; \cdots ; \mathbf{Q}_v]),\\& \mathbf{B} \in\{-1,1\}^{l \times n}
\end{aligned}
\label{eq.2}
\end{equation}
where kernelized bipartite graph $\phi{(\mathbf{X}_p)}\in \mathbf{R}^{m \times n}$ is the input data of the p-th view. $m$ and $n$  are the number of anchors and the number of samples. $\alpha$ is a hyperparameter used to adjust the proportion between the binary representation of multi-view data and the dual tensor. $\mathbf{Q}_{p}\in \mathbf{R}^{m \times l}$ is the projection matrix of the $p$-th view, used to project the bipartite graph into a consistent hamming space of dimensionality {$l$}. These projection matrices are stacked to form a tensor {$\tensor{Q}$}. {$\mathbf{B}_{p}\in \mathbf{R}^{l\times n}$} is the embedding feature matrix/hash matrix learned for the $p$-th view, and these embedding feature matrices are stacked to form a tensor {$\tensor{B}$}. {$\left| \middle| \cdot \middle| \middle| ~_{tnn} \right.$} denotes the Tensor Nuclear Norm (TNN), which could be solved by tensor Singular Value Decomposition ($t$-SVD) as mentioned in Fig. \ref{fig.t-svd}. In this way, TPCH captures the high-order semantic information among bipartite graphs and then embeds them into hash codes $\mathbf{B}_p\in\{-1,1\}$ to enhance expression ability.

\begin{figure}[h]
  \centering
  \includegraphics[width=\linewidth]{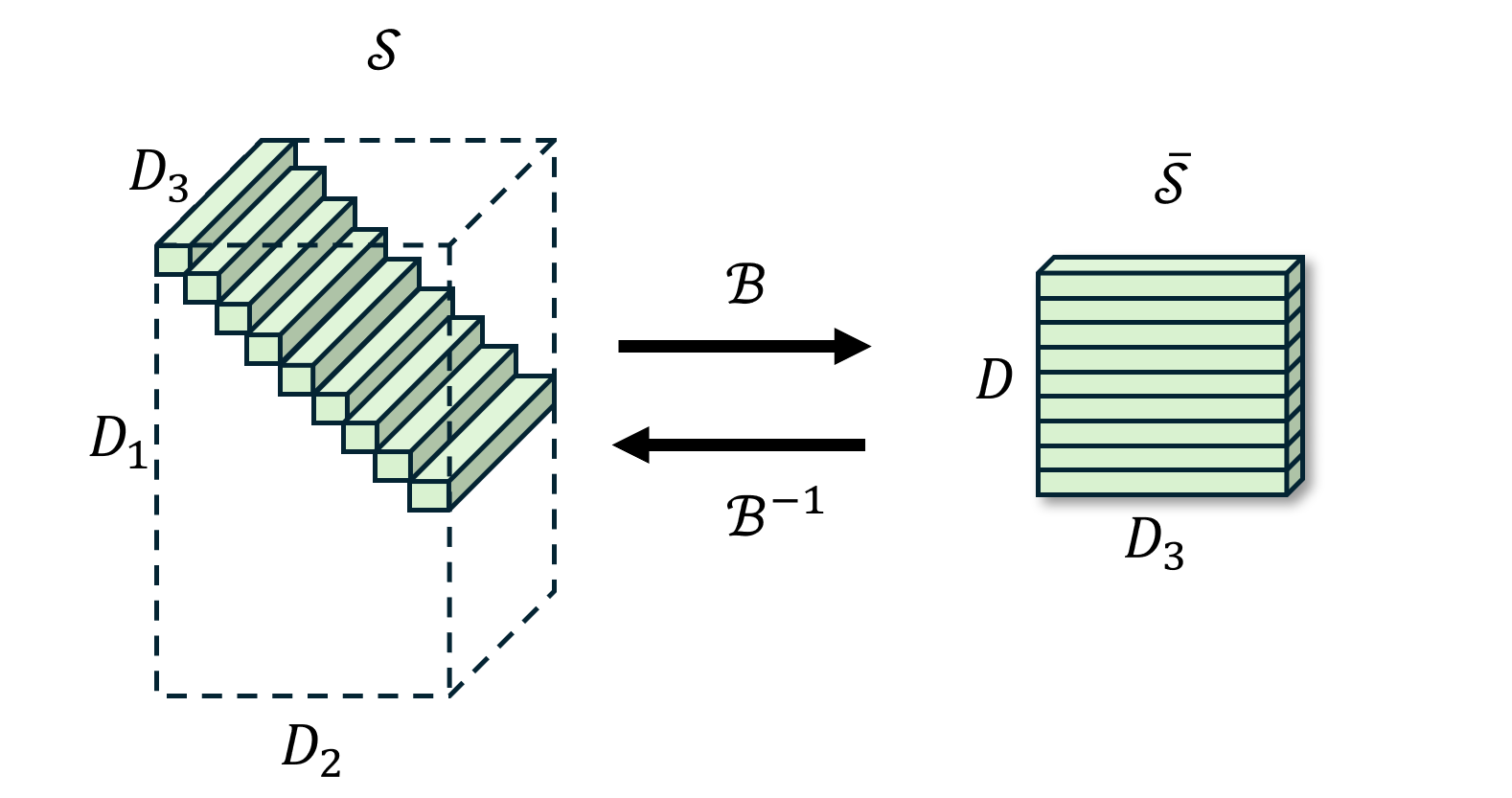}
  \caption{Transformation instructions for core tensors.}
  \label{fig.core-tensor}
\end{figure}

From Eq. \ref{eq.2}, TPCH could stack projection matrices and hash matrices to construct tensors $\tensor{Q}$ and $\tensor{B}$ to capture higher-order information of multi-view data in both dual projection and Hamming spaces. However, the tensors {$\tensor{Q}$} and {$\tensor{B}$} not only contain the desired high-order information but also incorporate noise or redundant information. As showcased in Fig. \ref{fig.t-svd}, the core tensor {$\tensor{S}$} from the t-SVD decomposition may not be low-rank, containing harmful noise and redundant information. Moreover, the core tensor with a high rank tremendously limits the higher-order synergistic between the projection matrices and hash codes. To reduce the impact of noise and redundant information, we develop the enhanced tensor nuclear norm by enforcing low-rank property on the core tensor as shown in Fig. \ref{fig.core-tensor}. 

By further ranking the core tensor in tensor Singular Value Decomposition ($t$-SVD) of the TNN, we further propose the Ehanced Tensor Nuclear Norm (ETNN) as
\begin{equation}
\begin{aligned}
\left| \left| \tensor{T} \right| \right|_{etnn} = \left| \middle| \overset{-}{\tensor{S}} \middle| \right|_{*} + \zeta\left| \middle| \tensor{U}\tensor{*}\tensor{B}^{- 1}\left( \overset{-}{\tensor{S}} \right)\tensor{*}\tensor{V} \middle| \right|_{*}
\end{aligned}
\label{eq.3}
\end{equation}
where {$~\overset{-}{\tensor{S}}{\in R}^{D_{~} \times D_{3}}$} is the low-rank approximation of the core tensor, {$\tensor{B}$} and {$\tensor{B}^{- 1}$} are transformations between {$\tensor{S}$} and {$\tensor{S}^{- 1}$}. {$\zeta$} is a predefined parameter, {$\|\cdot\|_{*}$} denotes the nuclear norm of the unfolded tensor. Fig. \ref{fig.core-tensor} illustrates the transformations between core tensors. Here, {$D = min\left( D_{1},D_{2} \right)$}, and {$\tensor{B}$} can be defined using matrix multiplication. By incorporating Fig. \ref{fig.core-tensor} into Fig. \ref{fig.t-svd}, the proposed ETNN could first alleviate the noise and redundant information hidden in the core tensor. Further, TPCH with ETNN ranks the core tensor in dual projection and Hamming spaces through transformations {$\tensor{B}$} and {$\tensor{B}^{- 1}$}, enhancing the robustness in the projection synergy and inter/intra-view communication. In this way, TPCH could generate more compact, distinguishable, and robust hash codes to improve the clustering performance. 

\subsection{Optimization}
For solving Eq. \eqref{eq.2} with ETNN, an alternating direction minimizing strategy is proposed to iteratively optimize our CHBG. Two auxiliary variables $\tensor{A}$ and $\tensor{E}$ are employed to get augmented Lagrangian function of Eq. \eqref{eq.2} as
\begin{equation}
\begin{aligned}
& \min_{\substack{\mathbf{B}_p, \mathbf{Q}_p,\tensor{Q},\\ \tensor{A},\tensor{B},\tensor{E}}}\alpha\sum_{p=1}^{v} \|\mathbf{Q}_p^\top \phi{(\mathbf{X}_p)} - \mathbf{B}_p\|_{F}^{2}+\frac{\mu}{2}\|\tensor{Q}-\tensor{A}+\frac{\tensor{Y}}{\mu}\|_{F}^{2} \\&+\|\tensor{A}\|_{etnn}+ \|\tensor{E}\|_{etnn}+\frac{\mu}{2}\|\tensor{B}-\tensor{E}+\frac{\tensor{J}}{\mu} \|_{F}^{2}
 \\& \text{s.t.}\ \mathbf{B} \in\{-1,1\}^{l \times n},\tensor{A} =\Phi([\mathbf{A}_1 ; \cdots ; \mathbf{A}_v]),\tensor{Q}=\tensor{A},\\&\tensor{B}=\tensor{E},
		\tensor{B} =\Phi([\mathbf{B}_1 ; \cdots ; \mathbf{B}_v]),\tensor{Q} =\Phi([\mathbf{Q}_1 ; \cdots ; \mathbf{Q}_v])
\end{aligned}
\label{eq.4}
\end{equation}
In this way, all the variables can be iteratively optimized one by one as follows.

\noindent \textbf{1- Updating the subproblem of  $\mathbf{Q}$:} Fixing the other variables of subproblem $\mathbf{Q}$, problem \eqref{eq.4} \wrt $\mathbf{Q}$ simplifies as
\begin{equation}
	\begin{aligned}
\alpha\sum_{p=1}^{v} \|\mathbf{Q}_p^\top \phi{(\mathbf{X}_p)} - \mathbf{B}_p\|_{F}^{2}	+\frac{\mu}{2}\|\mathbf{Q}_p-\mathbf{A}_p+\frac{\mathbf{Y}_p}{\mu}\|_{F}^{2}
\end{aligned}
	\label{eq.5}
\end{equation}
where $\mathbf{Q}_p= [2\alpha\mathbf{B}_p\phi^\top(\mathbf{X}_p ) +\mu(\mathbf{A}_p-\frac{\mathbf{Y}}{\mu} )]/[2\alpha\phi(\mathbf{X}_p)\phi^\top(\mathbf{X}_p) +\mu\mathbf{I}]$. $\mathbf{Q}$ subproblem requires $\mathcal{O}({m^3l+nml+nm^2})$ complexity.

\noindent \textbf{2- Updating the subproblem of  $\mathbf{B}$:} Fixing the other variables of subproblem $\mathbf{B}$, problem \eqref{eq.6} \wrt $\mathbf{B}$ simplifies as
\begin{equation}
\begin{aligned}
& \min _{\mathbf{B}_p}\alpha\sum_{p=1}^{v} \|\mathbf{Q}_p^\top \phi{(\mathbf{X}_p)} - \mathbf{B}_p\|_{F}^{2}+\frac{\mu}{2}\|\mathbf{B}_p-\mathbf{E}_p+\frac{\mathbf{J}_p}{\mu} \|_{F}^{2}
\\& \text { s.t. } \mathbf{B}_p \in\{-1,1\}^{l \times n}
\end{aligned}
	\label{eq.6}
\end{equation}
where 'con' is the constant part of $\mathbf{B}_p$. The constant $\operatorname{tr}\left(\mathbf{B}_p^\top \mathbf{B}_p\right)=\operatorname{tr}\left(\mathbf{B}_p \mathbf{B}_p^\top\right)=n l$ could make problem \eqref{eq.6} change to
\begin{equation}
\begin{aligned}
& \max_{\mathbf{B}_p} \sum_{p=1}^v \operatorname{tr}[\mathbf{B}_p^\top(\alpha\mathbf{Q}_p\phi{(\mathbf{X}_p)}+\frac{\mu}{2}(\mathbf{E}_p-\frac{\mathbf{J}_p}{\mu}))] \\
& \text { s.t. } \mathbf{B}_p \in\{-1,1\}^{l \times n}
\end{aligned}
	\label{eq.7}
\end{equation}
which has a closed-form optimal solution:
\begin{equation}
\mathbf{B}_p=\operatorname{sgn}(\alpha\mathbf{Q}_p\phi{(\mathbf{X}_p)}+\frac{\mu}{2}(\mathbf{E}_p-\frac{\mathbf{J}_p}{\mu}) .
	\label{eq.8}
\end{equation}
$\mathbf{B}$ subproblem requires $\mathcal{O}({mln+ncl})$ time complexity.

\noindent \textbf{Update-3: Solving $\tensor{A}$} can be written as
\begin{equation}
\begin{aligned}
\end{aligned}
\label{eq.9}
\end{equation}
which is solved by following steps \cite{lu2019tensor}: (i) minimizing the core matrix, and (ii) minimizing TNN.

\noindent(i) Updating core matrix as
\begin{equation}
\begin{aligned}
\min _{\bm{\mathfrak{P}}(\tensor{S})} \|\bm{\mathfrak{P}}(\tensor{S})\|_*+\frac{1}{2\lambda}\|\bm{\tensor{F}}-(\tensor{Z}+\frac{\tensor{Y}}{\mu}) \|_F^2
\end{aligned}
\label{eq.10}
\end{equation}
where regularization parameter $\lambda=1/(max(m,v)n)^{\frac{1}{2}}$. And the tensor $\tensor{S}$ is obtained from $t$-SVD on the temporary variable $\tensor{F}$, \ie $\tensor{F}=\tensor{U}*\tensor{S}*\tensor{V}$. 

\noindent (ii) Updating $\tensor{A}$ as
 \begin{equation}
\begin{aligned}
\min\limits_{\tensor{W}}\ \|\tensor{W}\|_{etnn} +\frac{\mu}{2}\|\tensor{W}-\tensor{G} \|_F^2 \\ 
\end{aligned}
\label{eq.11}
\end{equation}     

With the learned low-rank core matrix $\bm{\mathfrak{P}}(\hat{\mathbf{T}})$, we can use $t$-product to reconstruct a tensor as $\tensor{G}=\tensor{U}*\bm{\mathfrak{P}}^{-1}(\tensor{T})*\tensor{V}$. The learned $\tensor{G}$ can further produce a closed-form solution via \textbf{Theorem 1} provided in supplementary materials. Its complexity costs $\tensor{O}(ml+ml\log (m))$.

\textbf{Update-4: Solving $\tensor{E}$} is similar to the subproblem of $\tensor{A}$, which costs $\tensor{O}(nl+nl\log (n))$ complexity. To update the multiplier variable $\tensor{Y}$, we employ the following procedure:
\begin{equation}
	\begin{aligned}
		 &\tensor{Y}=\tensor{Y}+\mu(\tensor{Q}-\tensor{A}),\	 \eta=min(\rho_2\eta_2,\eta_{max})\\&\tensor{J}=\tensor{J}+\mu(\tensor{B}-\tensor{E}),\
		 \eta=min(\rho_3\eta_3,\eta_{max})
	\end{aligned}
	\label{eq.19}
\end{equation}
In this scenario, we set the parameter $\eta$ to $1e^{-4}$ and the upper bound for $\mu2$ ($\mu2_{max}$) to $10^{10}$. $\rho_2=2$ is employed. The computational complexity of the algorithm is linear to $n$. The solution of problem \eqref{eq.4} is provided in Algorithm \ref{alg.1}. $obj$ is the objective value.

After obtaining the multiple hash matrices $\{\mathbf{B}_p\}_{p=1}^v$, the final hashing matrix is further averaged as $\mathbf{\hat{B}}=\sum_{p=1}^v\mathbf{B}_p$. Then, the clustering indicator matrix could be achieved via the following 
\begin{equation}
\begin{aligned}
& \min _{\mathbf{C}, \mathbf{G}}\|\hat{\mathbf{B}}-\mathbf{C} \mathbf{G}\|_F^2 \\
& \text { s.t. } \mathbf{C}^\top 1=0, \mathbf{C} \in\{-1,1\}^{l \times k}, \mathbf{G} \in\{0,1\}^{k \times n}, \sum_i g_{i n}=1 .
\end{aligned}
	\label{eq.14}
\end{equation}
which can be solved according to the discrete proximal linearized minimization (DPLM) \cite{wang2023graph}.

Optimizing $\mathbf{G}$-Step is
\begin{equation}
\begin{aligned}
g_{i j}^{t+1}=\left\{\begin{array}{lr}
1, & j=\arg \min _* H\left(b_i, c_{,}^{t+1}\right), \\
0, & \text { otherwise, }
\end{array}\right.
\end{aligned}
	\label{eq.18}
\end{equation}
where $H\left(\mathbf{b}_i, \mathbf{c}^n\right)$ represents the Hamming distance between the $i$-th binary code $b_i$ and the $s$-th cluster centroid $c^s$, offering significantly faster computation compared to the Euclidean distance. The final results are obtained from $\mathbf{G}$.

\textbf{Time Complexity.} The main complexity cost of Algorithm \ref{alg.1} cost $\mathcal{O}( m^3l+nml+nm^2+mln+ncl+nl+nl\log (n)+ml+ml\log (m))$, which is linear to sample number $n$ since $\log (n)\ll n$.

\begin{algorithm}[!htpb]
  \caption{\  TPCH algorithm}
  \begin{algorithmic}[1]
  \REQUIRE  Bipartite graphs $\{\phi{(\mathbf{X}_p)}\}_{p=1}^v$ and Parameter $\alpha$.
  \ENSURE Clustering results.\\
  \REPEAT  
    \STATE  Update $\mathbf{Q}_p$ via Eq. \eqref{eq.5} ;\\
    \STATE  Update $\mathbf{B}_p$ via Eq. \eqref{eq.6} ;\\
        \STATE  Update $\tensor{A}$ via Eq. \eqref{eq.9} ;\\
                \STATE  Update $\tensor{E}$ similar to $\tensor{A}$;\\
        \UNTIL Satisfy convergence.\\
    \end{algorithmic}
    \label{alg.1} 
  \end{algorithm}

\begin{table*}[!tp]
    \centering
    \fontsize{9pt}{11pt}\selectfont
    \setlength{\tabcolsep}{2pt}
    \begin{tabular}{c|c|c|c|c|c|c|c|c|c|c|c|c|c|c|c|c|c|c|c|c}
        \toprule
        \multirow{2}{*}{Methods}  &\multicolumn{8}{c|}{Graph-based Methods} &\multicolumn{12}{c}{Hash-based Methods}\\
        \cmidrule[1pt]{2-21}
             & \text {SC} & \makecell{ Co-\\re-p } & \makecell{Co-\\re-c} & \makecell{AM\\GL} & \makecell{ Mul-\\NMF } & \makecell{ ML\\AN} & \text{ mPAC } & \text{GMC}  & \text { SH } & \text { DSH } & \text {SP} & \text { ITQ } & \text { SGH } & \makecell {RS\\SH} & \makecell{RF\\DH} & \makecell {HS\\IC} & \makecell{BM\\VC} & \makecell{AC-\\MVBC} & \makecell{GC\\AE} & \makecell{TP\\CH} \\

        \toprule
        & \multicolumn{8}{|c|}{\text{Acc}} & \multicolumn{12}{|c}{\text{Acc}} \\
        \toprule
        {\text { SUNRGBD }} & 0.11 & 0.18 & 0.18 & 0.10 & 0.14 & 0.19 & 0.13 & 0.22  & 0.12 & 0.16 & 0.19 & 0.19 & 0.18 & 0.16 & 0.17 & 0.16 & 0.14 & 0.19 & \textbf{0.24} & 0.22 \\    
        \hline
        {\text { Cifar-10 }} & 0.17 & 0.22 & 0.21 & 0.22 & 0.12 & 0.23 & 0.10 & 0.23  & 0.17 & 0.22 & 0.23 & 0.23 & 0.22 & 0.20 & 0.23 & 0.22 & 0.24 & 0.29 & 0.25 & \textbf{0.70}\\    
        \hline
        {\text { Caltech101 }} & 0.18 & 0.26 & 0.26 & 0.15 & 0.19 & 0.23 & 0.20 & 0.27  & 0.18 & 0.16 & 0.21 & 0.25 & 0.23 & 0.29 & 0.22 & 0.24 & 0.29 & 0.32 & 0.30 & \textbf{0.54}\\    
        \hline
        {\text { Caltech256 }} & 0.09 & 0.09 & 0.10 & 0.05 & 0.06 & 0.08 & 0.09 & 0.07  & 0.06 & 0.08 & 0.09 & 0.09 & 0.08 & 0.09 & 0.08 & 0.07 & 0.09 & 0.05 & 0.11 & \textbf{0.13}\\    
        \hline
        {\text { 100 leaves }} & 0.49 & 0.73 & 0.79 & 0.76 & 0.87 & 0.74 & 0.82 & 0.43 & 0.47 & 0.45 & 0.48 & 0.46 & 0.51 & 0.36 & 0.45 & 0.66 & 0.50 & 0.83 & \textbf{0.89}  & 0.86\\    
        
        \toprule
        & \multicolumn{8}{|c|}{\text{NMI}} & \multicolumn{12}{|c}{\text{NMI}} \\
        \toprule
        {\text { SUNRGBD }} & 0.01 & 0.21 & 0.22 & 0.19 & 0.10 & 0.21 & 0.07 & 0.22  & 0.13 & 0.22 & 0.22 & 0.21 & 0.22 & 0.20 & 0.20 & 0.22 & 0.16 & 0.24 & 0.22 & \textbf{0.28}\\    
        \hline
        {\text { Cifar-10 }} & 0.077 & 0.09 & 0.09 & 0.09 & 0.02 & 0.09 & 0.01 & 0.10  & 0.03 & 0.09 & 0.10 & 0.10 & 0.10 & 0.07 & 0.10 & 0.09 & 0.10 & 0.13 & 0.10 & \textbf{0.67}\\    
        \hline
        {\text { Caltech101 }} & 0.18 & 0.26 & 0.26 & 0.15 & 0.19 & 0.23 & 0.20 & 0.27  & 0.33 & 0.36 & 0.40 & 0.44 & 0.44 & 0.49 & 0.44 & 0.45 & 0.49 & 0.51 & 0.47 & \textbf{0.80}\\    
        \hline
        {\text { Caltech256 }} & 0.28 & 0.30 & 0.28 & 0.11 & 0.15 & 0.28 & 0.22 & 0.14  & 0.26 & 0.29 & 0.30 & 0.27 & 0.30 & 0.31 & 0.29 & 0.24 & 0.32 & 0.22 & 0.29 & \textbf{0.37}\\    
        \hline
        {\text { 100 leaves }} & 0.77 & 0.88 & 0.93 & 0.91 & 0.93 & 0.89 & 0.93 & 0.70  & 0.72 & 0.73 & 0.72 & 0.74 & 0.76 & 0.62 & 0.72 & 0.83 & 0.73 & 0.93 & 0.94 & \textbf{0.97}\\    
        
        \toprule
        & \multicolumn{8}{|c|}{\text{Purity}} & \multicolumn{12}{|c}{\text{Purity}} \\
        \toprule
        {\text { SUNRGBD }} & 0.11 & 0.33 & 0.34 & 0.11 & 0.16 & 0.33 & 0.14 & 0.23  & 0.24 & 0.33 & 0.34 & 0.34 & 0.34 & 0.33 & 0.33 & 0.35 & 0.28 & 0.34 & 0.34 & \textbf{0.37}\\    
        \hline
        {\text { Cifar-10 }} & 0.18 & 0.22 & 0.22 & 0.23 & 0.12 & 0.25 & 0.10 & 0.25  & 0.17 & 0.23 & 0.23 & 0.23 & 0.22 & 0.21 & 0.23 & 0.22 & 0.24 & 0.27 & 0.26 & \textbf{0.73}\\    
        \hline
        {\text { Caltech101 }} & 0.31 & 0.46 & 0.47 & 0.17 & 0.32 & 0.44 & 0.30 & 0.35  & 0.31 & 0.34 & 0.39 & 0.42 & 0.41 & 0.47 & 0.42 & 0.41 & 0.49 & 0.46 & 0.44 & \textbf{0.70}\\    
        \hline
        {\text { Caltech256 }}  & 0.13 & 0.14 & 0.15 & 0.04 & 0.11 & 0.14 & 0.14 & 0.10  & 0.11 & 0.13 & 0.14 & 0.14 & 0.14 & 0.15 & 0.13 & 0.11 & 0.15 & 0.08 & 0.14 & \textbf{0.18}\\    
        \hline
        {\text { 100 leaves }} & 0.56 & 0.76 & 0.83 & 0.81 & 0.90 & 0.76 & 0.85 & 0.54  & 0.50 & 0.50 & 0.51 & 0.49 & 0.53 & 0.39 & 0.48 & 0.68 & 0.53 & 0.88 & \textbf{0.90} & 0.88\\    
        
        \bottomrule        
    \end{tabular}
    \caption{Comparison results with graph-based and hash-based methods}
    \label{tab.1}
\end{table*}

\section{Experiment}
In this section, we introduce the datasets, comparisons, evaluation metrics, and experiment analysis. To assess the performance of TPCH, we compare two categories of competitors, \ie graph-based methods and hash-based methods. Additionally, we further evaluate parameter sensitivity analysis, convergence analysis, and ablation studies.

\subsection{Experimental Setting}
\subsubsection{Five benchmark multi-view datasets include:}
\begin{itemize}
    \item SUNRGBD: This dataset consists of colour images and depth images of indoor scenes to provide rich information. This dataset comprises 37 categories, totalling 10,335 images.
    \item Cifar-10: This dataset consists of colour images categorized into 10 different classes, each containing 6,000 32x32-pixel colour images.
    \item Caltech101: Created by the California Institute of Technology, this dataset contains 101 object categories with approximately 9,000 images, including animals, plants, vehicles, and various other objects.
    \item Caltech256: Similar to Caltech101, it comprises 256 categories, totaling approximately 30,607 images.
    \item 100 leaves: This dataset includes 100 plant species with a total of 1,600 samples. Each sample consists of three views: shape, fine-scale margin, and texture histograms.
\end{itemize}

\subsubsection{Comparison methods and setting:}
To evaluate the validity of our TPCH, two categories of competitors are employed: (1) Hash-based methods include SH \cite{weiss2008spectral}, DSH \cite{jin2013density}, SP \cite{xia2015sparse}, ITQ \cite{gong2012iterative}, SGH \cite{jiang2015scalable}, RSSH \cite{tian_unsupervised_2020}, RFDH \cite{wang2017robust}, HSIC \cite{zhang2018highly}, BMVC \cite{zhang2018binary}, AC-MVBC \cite{zhang2022learning} and GCAE \cite{wang2023graph}. (2) Graph based methods include SC \cite{ng2001spectral}, Co-regularize \cite{kumar2011co}, AMGL \cite{nie2016parameter}, Mul-NMF \cite{liu2013multi}, MLAN \cite{nie2017multi}, mPAC \cite{kang2019multiple}, GMC \cite{wang2019gmc} and GCAE. For these methods, we conduct experiments using  publicly available source codes and parameter settings of the competitors. By comparing against a large number of advanced methods, we can thoroughly verify the effectiveness of our TPCH. In the experiment, five mainstream metrics are used to evaluate clustering performance: Accuracy (ACC), Normalized Mutual Information (NMI), Purity, F-score, and Adjusted Rand Index (ARI). Due to space limitations, their definitions, along with the experimental results of F-score and ARI, are provided in the supplementary materials. The hardware for running our experiments includes CPU: i9  14900k 24-core processor, RAM: 32GB 7200MHz, Platform: Windows 11, Software: MATLAB 2023b.

\subsection{Comparative experimental results}
Table~\ref{tab.1} reports the clustering performance of TPCH and various hash-based 
 and graph-based clustering methods on five multi-view datasets, with the optimal results highlighted in bold. According to the results of the five clustering evaluation metrics, Table 1 demonstrates that:
\begin{itemize}
    \item AC-MVBC, GCAE, and our TPCH outperform other methods significantly. The reasons may be that: 1) AC-MVBC and our TPCH could capture the high-order information among multi-view data, which enhances the discriminability of hash codes; 2) GCAE and our TPCH encode the dynamic graph semantic information into hash codes to enlarge their quality. 
   \item Compared to most recently GCAE and AC-MVBC, TPCH also outperforms them consistently in most cases. The main reasons are that 1) our TPCH first captures the high-order information of data in dual projection and Hamming spaces to learn the more compact and discriminable hash representations as shown in Fig. \ref{fig.tsne_competitor} and Fig. \ref{fig.tsne_iter}. This phenomenon indicates that our enhanced tensor nuclear norm on dual projection and Hamming spaces could simultaneously promote the projection synergy, as well as inter-view and intra-view communication. 
   \item Conventional multi-view clustering methods use Euclidean distance to measure the distance between two samples, which leads to inefficiency and high computational complexity. In contrast, hash-based methods obtain clustering results in the Hamming space, thereby improving computational efficiency.
\end{itemize}

\begin{figure}[htbp]
  \centering
  \subfigure[Caltech101]{\includegraphics[width=0.47\linewidth]{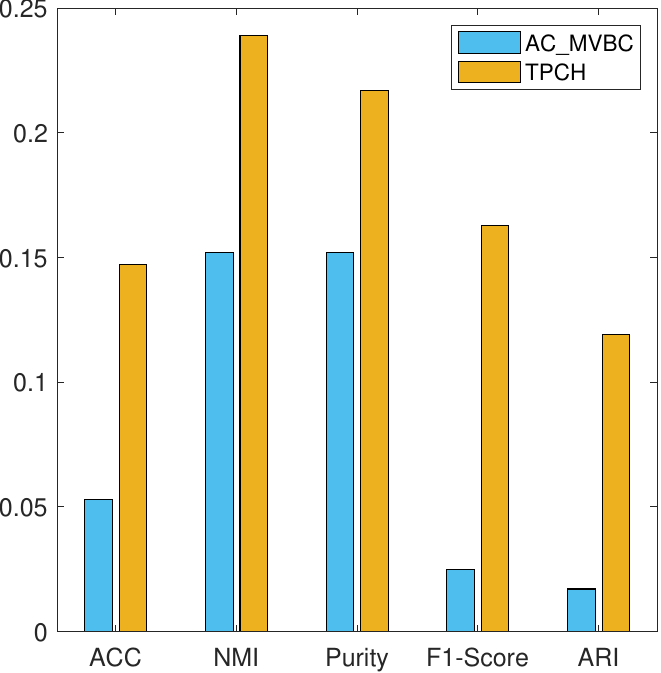}}
  \subfigure[Caltech256]{\includegraphics[width=0.47\linewidth]{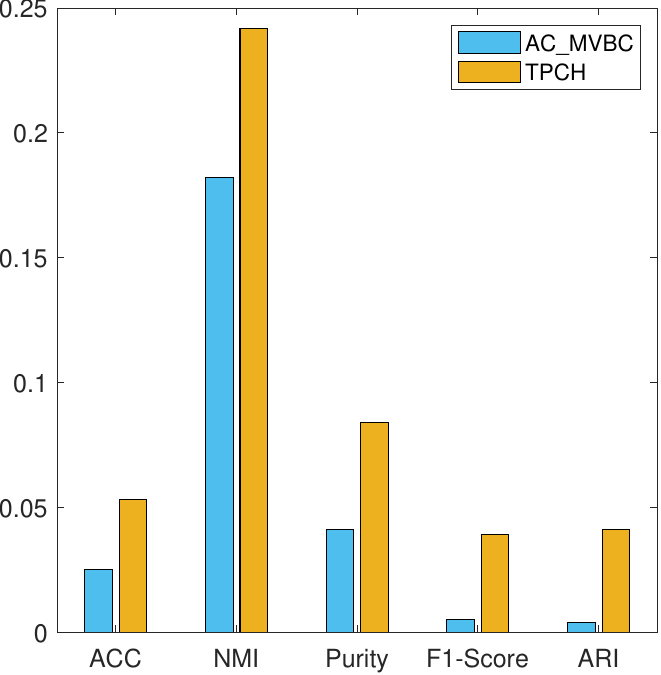}}
  \caption{Clustering performance of AC\_MVBC and TPCH on the two datasets with salt-and-pepper noise.}
\label{fig.robust_analysis}
\end{figure}

\noindent \textbf{Robust analysis:} Fig. \ref{fig.robust_analysis} reports the comparison experiment results of TPCH with AC-MVBC on the Caltech101 and Caltech256 datasets with salt-and-pepper noise. Experiments conducted on the SUNRGBD, Cifar-10 and 100 leaves datasets are provided in the Supplementary Materials. Through Fig. \ref{fig.robust_analysis}, it can be observed that TPCH is completely superior to AC-MVBC on five main clustering indicators. Similar to TPCH, AC-MVBC also uses tensor nuclear norm to capture the latent higher-order correlations in multi-view data, but it ignores the low-rank property of the core tensor, making existing tensor nuclear norm more susceptible to noise. More importantly, core tensor with high rank limits the higher-order synergistic between the projection matrices. By further explore the low-rank property of core tensor on dual projection and Hamming spaces, the projection cooperation and inter-view/intra-view communication would negotiate with each other to achieve the compact hash representation. Thus, our TPCH could 
enjoy more robustness to noise and better generate the more compact hash codes for improve the clustering performance· Furthermore, experiments on noisy datasets also verify the robustness of TPCH to noise as mentioned in Fig. \ref{fig.noise_analysis}.

Table~\ref{tab.time} demonstrates the runtime of TPCH compared to the state-of-the-art method, GCAE, across five large-scale datasets. It is evident from Table~\ref{tab.time} that TPCH significantly outperforms GCAE in terms of computational speed despite both methods being based on hashing. The time complexity of GCAE is $\mathcal{O}(n^3+3n^2l+n^2 +nlc)$, which implies cubic complexity with respect to the number of samples, whereas TPCH has a time complexity of $\mathcal{O}( m^3l+nml+nm^2+mln+ncl+nl+nl\log (n)+ml+ml\log (m))$, indicating linear complexity for the number of samples. The lower time complexity grants TPCH higher computational efficiency. 

In summary, TPCH not only excels in various metrics for executing multi-view clustering tasks but also significantly surpasses existing algorithms in computational efficiency, making it well-suited for rapidly processing large-scale multi-view clustering tasks.

\begin{table}[H]
	\centering
	\footnotesize
  \begin{tabular}{c|c|c|c}
			\toprule
			\text { Datasets }   & \text { GCAE } & \text { TPCH } & \text { Speedup }  \\
			\toprule
			\multicolumn{4}{c}{\text{CPU-Time(s)}} \\
			\toprule
			{\text { SUNRGBD }}  &1188.33 &157.62 & \textbf{7.54$\times$} \\	
			\hline
			{\text { Cifar-10 }} &1120.15 &157.25 & \textbf{7.12$\times$} \\	
			\hline
			{\text { Caltech101 }} &1158.71 &114.11 & \textbf{10.15$\times$} \\	
			\hline
			{\text { Caltech256 }}  &47618.00 &5406.76  & \textbf{8.81$\times$} \\	
			\hline
			{\text { 100 leaves }}  &14.16 &1.75  & \textbf{8.09$\times$} \\	
   
			\bottomrule		
		\end{tabular}%
            \caption{Running time of GCAE and TPCH.}
		\label{tab.time}%
	\end{table}%

\subsection{Ablation Studies}
\begin{figure}[htbp]
  \centering
  \subfigure[Caltech101]{\includegraphics[width=0.47\linewidth]{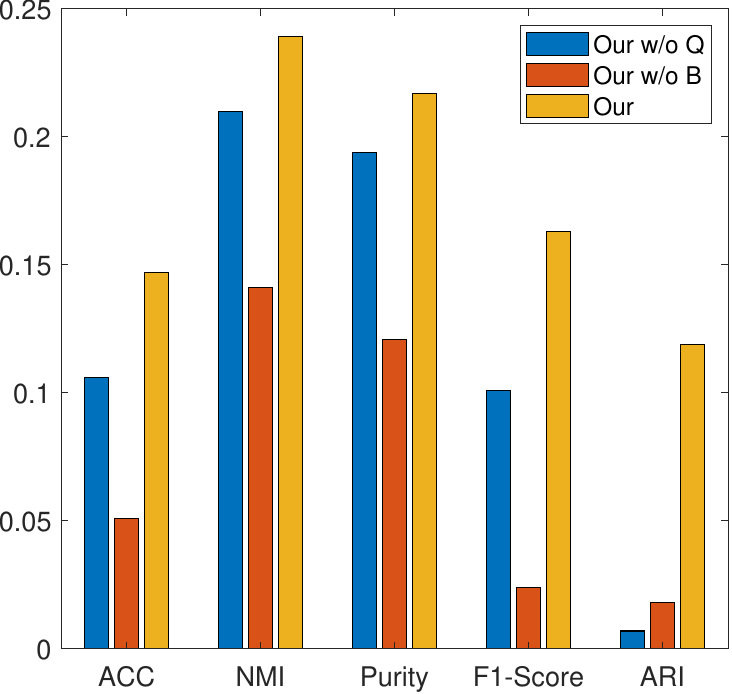}}
  \subfigure[Caltech256]{\includegraphics[width=0.47\linewidth]{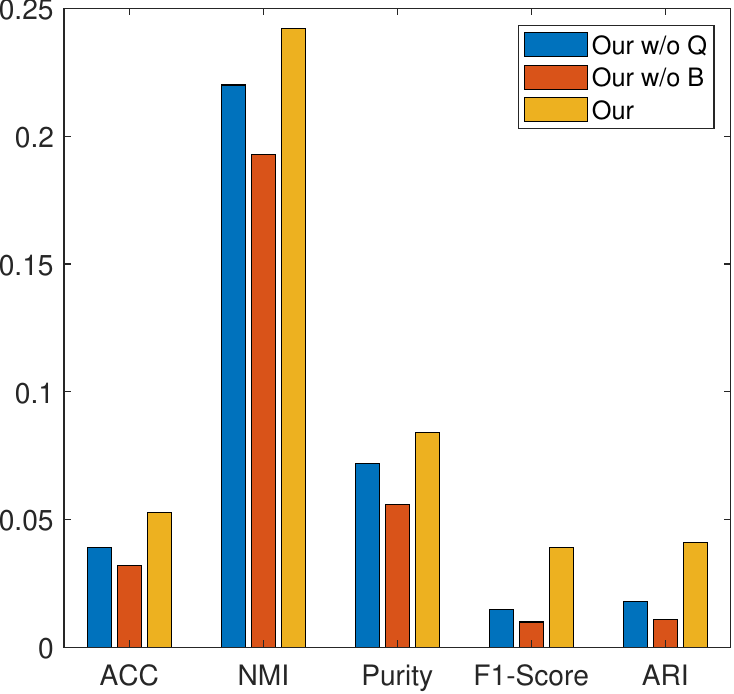}}
  \caption{Ablation Studies of TPCH on the Caltech101 and Caltech256 datasets with salt-and-pepper noise.}
\label{fig.noise_analysis}
\end{figure}

The TPCH model mainly includes two tensors:  \textbf{$\tensor{Q}$} and \textbf{$\tensor{B}$}. To further investigate why TPCH performs well, we remove \textbf{$\tensor{Q}$} and \textbf{$\tensor{B}$} respectively, naming these methods as without (w/o) \textbf{$\tensor{B}$} and without (w/o) \textbf{$\tensor{B}$}, while conducting experiments with datasets processed with salt-and-pepper noise. The results of the ablation study are shown in Fig.\ref{fig.noise_analysis}. From Fig. \ref{fig.noise_analysis}, we observe that the removal of \textbf{$\tensor{Q}$} and \textbf{$\tensor{B}$} leads to a significant decline in clustering performance, verifying the robustness of proposed enhanced tensor nuclear norm on projection matrices and hash codes. 

\subsection{Visualization of Clustering Results}
\begin{figure}[h]
  \centering
  \subfigure[AC-MVBC]{\includegraphics[width=0.45\linewidth]{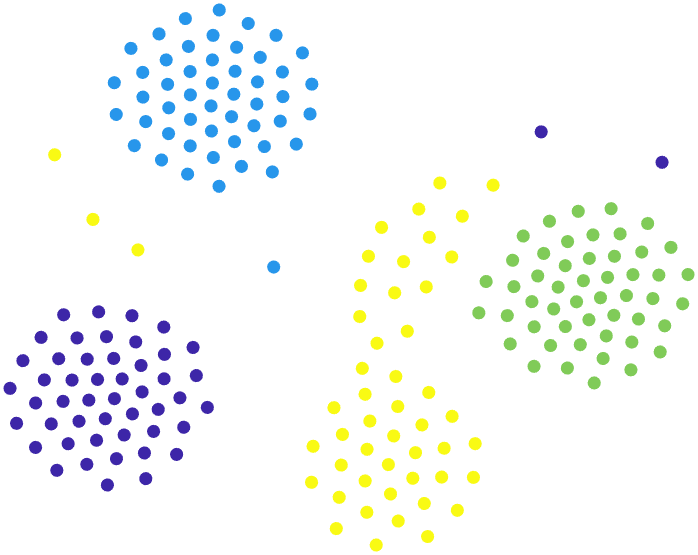}\label{fig.tsne_competitor:suba}}\quad
  \subfigure[TPCH]{\includegraphics[width=0.45\linewidth]{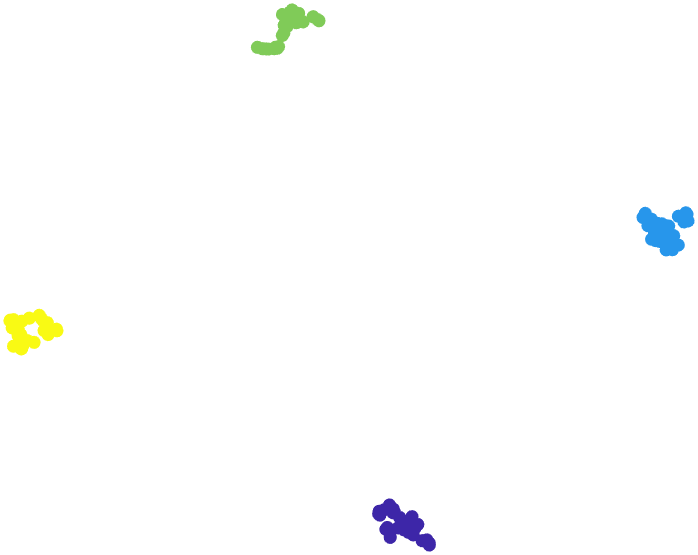}\label{fig.tsne_competitor:subb}}
  \caption{Visualization of clustering results for TPCH and AC-MVBC on the Synthetic\_4clu dataset.}
\label{fig.tsne_competitor}
\end{figure}

The visualization of clustering results for AC-MVBC and TPCH on the Synthetic\_4clu datasets is presented in Fig.\ref{fig.tsne_competitor}. Different colors represent different clusters learned on the hash codes. We also visualized the clustering results on the Synthetic\_3d dataset. Due to space constraints, we have included it in the supplementary materials. 

Fig. \ref{fig.tsne_competitor:suba} and Fig. \ref{fig.tsne_competitor:subb} reveals that the clustering performance of TPCH on the Synthetic\_4clu dataset is significantly better than AC-MVBC. Fig. \ref{fig.tsne_competitor:subb} exhibits larger inter-cluster distances and smaller intra-cluster distances. Additionally, the clusters displayed in Fig. \ref{fig.tsne_competitor:subb} are more compact, directly demonstrating the effectiveness of TPCH in improving hash code compactness.

\begin{figure}[h]
  \centering
  \subfigure[Synthetic\_4clu:iter=2]{\includegraphics[width=0.45\linewidth]{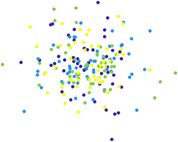}\label{fig.tsne_iter:a}}\quad
  \subfigure[Synthetic\_4clu:iter=22]{\includegraphics[width=0.45\linewidth]{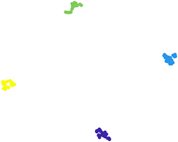}\label{fig.tsne_iter:b}}\\
    \subfigure[Synthtic\_3d:iter=2]{\includegraphics[width=0.45\linewidth]{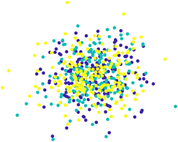}\label{fig.tsne_iter:c}}\quad
  \subfigure[Synthtic\_3d:iter=20]{\includegraphics[width=0.45\linewidth]{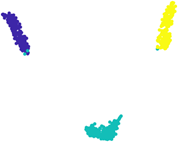}\label{fig.tsne_iter:d}}
  \caption{Clustering performance of TPCH at the beginning and after convergence of clustering on the Synthetic\_4clu and Synthetic\_3d datasets.}
\label{fig.tsne_iter}
\end{figure}

Fig. \ref{fig.tsne_iter} displays the performance of TPCH at the beginning and after convergence of clustering on the Synthetic\_4clu and Synthetic\_3d datasets. It can be seen that in both datasets, TPCH exhibits a clear clustering effect after convergence. It is noteworthy that, despite the more complex data distribution in Synthetic\_3d compared to Synthetic\_4clu, TPCH can still effectively partition the clusters. The stable clustering performance of TPCH under these two different initial data distributions once again proves the superiority of TPCH.

\begin{figure}[t]
  \centering
\subfigure[]{\includegraphics[width=0.48\linewidth]{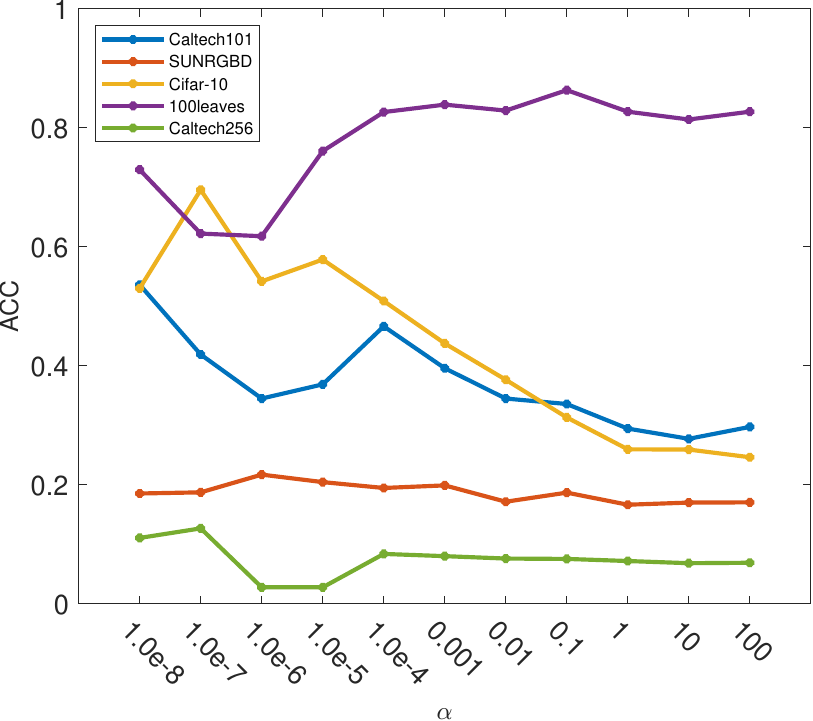}}
\subfigure[]{\includegraphics[width=0.48\linewidth]{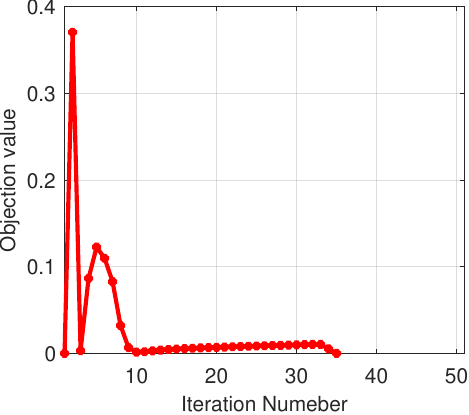}}
  \caption{Sensitive analysis \wrt $\alpha$ for five large-scale datasets and convergence curve for the Caltech256 dataset.}
  \label{fig.senstive}
\end{figure}

\subsection{Sensitivity and Convergence Analysis}
In TPCH, we set the dimensionality of the hash codes and the representation space to be 64. TPCH has only one hyperparameter, \textbf{$\alpha$}, which adjusts the balance between the binary representation of multi-view data and tensor factorization. We employ a logarithmic search approach, varying \textbf{$\alpha$} between 1.0e-8 and 100 at each order of magnitude. Fig. \ref{fig.senstive} illustrates the sensitivity of different datasets to \textbf{$\alpha$} across the clustering metrics ACC. The 100leaves dataset maintains high performance across all \textbf{$\alpha$}. As shown in Fig. \ref{fig.senstive} (a), TPCH achieves satisfactory results over a wide range of \textbf{$\alpha$} values, indicating insensitivity to parameters. For the sensitivity analysis of more metrics, please refer to 
supplementary materials. Fig. \ref{fig.senstive} (b) illustrates the convergence curves of TPCH on the Caltech256 datasets, which demonstrates that TPCH converges very quickly. For the convergence of more metrics, please refer to supplementary materials.


\section{Conclusion}
In conclusion, this work introduces a novel Tensor-interacted Projection and Cooperative Hashing for Multi-view Clustering(TPCH) approach for multi-view clustering, which effectively addresses the limitations of existing methods by considering the synergistic interactions and communications during the projection process. This innovative approach significantly enhances the compactness and distinguishability of hash representations, thereby improving both the quality of clustering and the robustness against noise. Experimental results validate the effectiveness of TPCH in multi-view clustering scenarios. 

\section{Acknowledgments}
This work is supported by the National Natural Science Foundation of China (project no. 62406069), the National Natural Science Foundation of China (Grant No. 62372235), the China Postdoctoral Science Foundation (project no. 2024M750425), the Mianyang Science and Technology Program (Grant nos. 2023ZYDF003, 2023ZYDF091), the Sichuan Science and Technology Program (Grant no. 2023YFG0135), and the Sichuan Science and Technology Program (Grant No. MZGC20240057).

\bibliography{aaai25}

\vfill
\hfill

\end{document}